\pdfoutput=1  
\documentclass[12pt]{article}

\usepackage[margin=1in]{geometry}
\usepackage{amsmath,amssymb}
\usepackage{mathptmx}   
\usepackage{graphicx}
\usepackage{booktabs}
\usepackage{caption}
\usepackage{subcaption}
\usepackage{siunitx}
\usepackage[hidelinks]{hyperref}
\usepackage{authblk}
\usepackage{xcolor}
\usepackage{placeins}
\usepackage{float}      

\graphicspath{{semi_final_images_v3/scaling/}{semi_final_images_v3/data_scaling/}{semi_final_images_v3/seqlen_scaling/}{semi_final_images_v3/sweep/}}

\title{\textbf{The Von--Neumann State--Space Transformer for Neural Decoding}}
\author{Morteza\ Sarafyazd\thanks{Affiliation: BrainCo, Somerville, US.}\\ \raisebox{15pt}[0pt][0pt]{\normalsize\texttt{morteza.sarafyazd@brainco.tech}}}
\date{}

\begin{document}
\maketitle

\begin{abstract}
Cortical computation is strikingly low--dimensional: a handful of latent variables,
carried in a neural population's activity, steer the higher--dimensional
responses of individual neurons. Our aim is sample efficiency---models
that decode well from limited data and at small parameter budgets. In a standard
Transformer layer, the feed--forward block applies the same operator to every token.
We suggest a von--Neumann inspired hypothesis of efficient computation as an alternative
for neural decoding: a
controller decodes an instruction and then executes a token--specific operator;
the usual realization---a soft mixture of experts---only blends their outputs, not
operators. We introduce a Von--Neumann State--Space
Transformer (VN--SST), a
memory--augmented Transformer whose feed--forward block is a low--rank instruction
bank: a shared base operator plus a small set of learned low--rank instructions, from
which a per--token code synthesizes the weight matrix actually used at that token. The
code is read from a low--dimensional projection of a carried state--space memory, so a
slow latent trajectory acts as an instruction pointer---mirroring how low--dimensional
dynamics may route cortical computation. On three motor--cortex
neural--decoding benchmarks, VN--SST is far more data--efficient than a modern
Transformer, each jointly predicting spikes and decoding behavior. This model
wins by a wide margin on the scarcest benchmark, leads on the other two, and turns longer
context into rising rather than falling accuracy. We evaluated that
the network compresses a large instruction bank to a few
bits per token, so program capacity acts as a control channel, not an
accuracy lever. The same model is also more parameter--efficient on two small
text benchmarks used for language modeling (LLMs), suggesting a generic mechanism.
\end{abstract}

\section{Introduction}
Cortical computation is strikingly low--dimensional: population activity spanning
thousands of neurons is organized by a handful of latent variables, whose slow
trajectories steer the faster, higher--dimensional responses of individual cells. Our
focus is neural decoding that is sample--efficient---accurate behavioral read--out from
limited data and small models. To pursue it
we adopt a von--Neumann--inspired hypothesis of efficient computation and build it into a
Transformer backbone. A von--Neumann (stored--program) machine has the same structure: a
small, slow controller fetches and decodes an instruction, and an
execution unit then runs the operator that instruction names, so a compact program steers
a large, input--specific computation. In a standard Transformer, by contrast, the
feed--forward network (FFN) applies the same fixed weights to every token; we instead give
the FFN a controller---a slowly varying state--space memory whose read--out selects, token
by token, the operator the FFN executes.

A natural first attempt is a soft mixture of experts (soft--MoE), in which a controller
blends the outputs of several fixed FFN experts~\cite{shazeer2017moe}. This is a useful
capacity knob, but in the von--Neumann sense it does not execute a program: it mixes the
results of fixed operators rather than constructing a token--specific one. We instead let
the controller synthesize the FFN's weights themselves, per token, from a small shared
bank of low--rank ``instructions'' added to a shared base operator (Eq.~\ref{eq:bank},
Section~\ref{sec:model3}). Each token's code selects a point on a low--dimensional manifold
of operators, so the executed map is genuinely token--specific while each added instruction
costs little---exposing the size of the instruction bank as a scaling axis of its own,
alongside parameters and data, and turning the program into a concrete control signal we
can measure.

The program is generated rather than looked up. We do not read the instruction off the raw
token; instead we condition it on a low--dimensional read--out of a slowly varying
selective--state--space (SSM) memory carried across the sequence. The slow latent trajectory then acts as an
instruction pointer---low--dimensional dynamics that fetch which operator to run---turning
the neuroscience intuition into a mechanism: a small, slow latent drives the
higher--complexity, token--specific computation.

We realize this in VN--SST, a Von--Neumann State--Space Transformer: a drop--in,
memory--augmented Transformer---local attention, a selective SSM, and a fast--weight
memory, coordinated by a controller---whose feed--forward operator is synthesized per token
from the low--rank instruction bank driven by the carried low--dimensional state
(Section~\ref{sec:model3}). As in a scaling--law
study~\cite{kaplan2020scaling}, we compare VN--SST against a modern Transformer on three
motor--cortex decoding codecs along three axes---parameters, data, and context
(Sections~\ref{sec:param}--\ref{sec:seqlen}). The programmable operator is markedly more
sample--efficient: under a limited--data budget it beats the Transformer by a wide margin on
the scarcest codec (decode $R^2$ $0.35$ vs.\ $0.21$), leads at every data budget on all
three, and---uniquely---turns a longer context window into rising rather than falling
decode accuracy. A control--bits diagnostic (Section~\ref{sec:controlbits}) then makes the
von--Neumann claim measurable: the network compresses a $32$--instruction bank to only a few
bits per token (${\approx}3.4$--$6.5$ effective operators), so program capacity behaves as a
control channel, not an accuracy lever---a signal no output--blending mixture can report.

\section{Models}
\label{sec:models}
Both models share a task--agnostic backbone mapping a hidden sequence
$h\in\mathbb{R}^{B\times T\times d}$ (batch size $B$, sequence length $T$, and hidden
width $d$) to an output of the same shape; only the
input/output heads differ. For the neural codec the input head is a linear map from
the $N$ binned firing rates and there are two output heads: a linear map back to $N$
rates (spike continuation) and a linear behavioral read--out.

\subsection{Transformer baseline}
The baseline is a modern decoder--only Transformer~\cite{vaswani2017,touvron2023llama}: a stack of
identical layers, each combining pre--norm RMSNorm~\cite{zhang2019rmsnorm}, rotary
position embeddings (RoPE)~\cite{su2021roformer}, causal multi--head self--attention, and a
SwiGLU~\cite{shazeer2020glu} feed--forward block. Every layer applies attention and the
feed--forward block as two residual updates,
\begin{equation}
h \leftarrow h + \mathrm{Attn}\!\big(\mathrm{RMSNorm}(h)\big),
\qquad
h \leftarrow h + \mathrm{SwiGLU}\!\big(\mathrm{RMSNorm}(h)\big),
\label{eq:tflayer}
\end{equation}
where the feed--forward block gates one linear projection of its input by another,
\begin{equation}
\mathrm{SwiGLU}(x)=W_2\big(\mathrm{SiLU}(W_1x)\odot W_3x\big).
\label{eq:swiglu}
\end{equation}
The three weight matrices $W_1,W_3,W_2$ are shared across positions, so the same operator
acts on every token.

\subsection{VN--SST: memory, state space, and programmable computation}
\label{sec:model3}
VN--SST keeps the layer skeleton of the baseline but replaces its two blocks: three
parallel memory pathways in place of plain attention, and a programmable feed--forward
operator in place of the fixed SwiGLU. These choices are a direct reading of the stored--program hypothesis that motivates this
work. A von--Neumann machine separates a controller that fetches and decodes an instruction,
an execution unit that runs the decoded operator, and a memory that persists state between
steps; computation is the loop that reads the next instruction and applies it. VN--SST maps
these roles onto a single sequence layer: the persistent selective state--space and
fast--weight pathways are the memory, carrying state across tokens; a low--dimensional
read--out of the slow state serves as the instruction pointer that selects which operator to
run next; the controller decodes that pointer into a per--token instruction code; and the
programmable SwiGLU is the execution unit, whose weights that code synthesizes on the fly.
The three pathways and the low--rank instruction bank described below are these roles made
concrete.

Each layer first normalizes its input,
$u=\mathrm{RMSNorm}(h)$, and a small controller reads off per--token gates---read--mix gates
$g_t\in\Delta^2$ (a simplex weighting over the three pathways) and memory write gates
$w^{\Delta}_t,w^{M}_t=\sigma(\cdot)$. The three pathways then run in parallel.

\paragraph{Pathway 1 --- local sensory buffer.}
The first pathway is ordinary multi--head self--attention restricted to a causal band of
width $w$,
\begin{equation}
y^{\text{loc}}=\mathrm{LocalAttn}_w(u),
\label{eq:local}
\end{equation}
which costs only $\mathcal{O}(Tw)$ and captures short--range structure inside the current
window, but carries nothing across windows---that is the job of the two persistent pathways
that follow.

\paragraph{Pathway 2 --- selective state space (slow dynamics).}
The second pathway is a diagonal, input--dependent SSM~\cite{gu2022s4,gu2023mamba} that
carries a per--channel state $s\in\mathbb{R}^{d\times n}$ ($n$ latent states per channel)
across segments. With input projection $x=W_xu$, per--token step size
$\Delta_t=w^{\Delta}_t\,\mathrm{softplus}(W_\Delta u_t+b)$ (write gate $w^{\Delta}_t$;
learned $W_\Delta$ and bias $b$), input--dependent selection vectors
$B_t,C_t\in\mathbb{R}^n$, and diagonal decay $A=-\exp(A_{\log})$ (learned $A_{\log}$),
\begin{align}
\bar A_t=\exp(\Delta_t\odot A), \quad
s_t=\bar A_t\odot s_{t-1}+(\Delta_t x_t)\otimes B_t, \quad
y^{\text{ssm}}_t=\langle s_t,C_t\rangle+D\odot x_t .
\label{eq:ssm}
\end{align}
Here $\odot$ is the elementwise (Hadamard) product, $\otimes$ the outer product,
$\langle\cdot,\cdot\rangle$ contracts over the $n$ state dimensions, and $D$ is a learned
per--channel skip. The read--out $y^{\text{ssm}}_t$ is a low--dimensional projection of the
slow state $s_t$; it is exactly the signal we use as the instruction pointer below.

\paragraph{Pathway 3 --- fast--weight associative memory (episodic).}
The third pathway is a delta--rule matrix memory $M\in\mathbb{R}^{d_k\times d_v}$ (key and
value dimensions $d_k,d_v$), also carried across segments, with $\ell_2$--normalized keys
and queries $k_t,q_t\in\mathbb{R}^{d_k}$ and a value $v_t\in\mathbb{R}^{d_v}$ (write gate
$w^{M}_t$). At each step it writes the current prediction error into $M$ and reads content
back by query through an output projection $W_o$,
\begin{align}
M_t=M_{t-1}+w^{M}_t\,k_t\big(v_t-k_t^{\top}M_{t-1}\big)^{\top}, \quad
y^{\text{mem}}_t=W_o\big(q_t^{\top}M_t\big).
\label{eq:mem}
\end{align}
Writing the error rather than the raw value makes recall content--addressable: a key that
resembles a stored one retrieves its associated value.

\paragraph{Fusion.}
The three read--outs are blended by the controller's read gates and added back to the
residual stream through a linear map $W_f$,
\begin{equation}
r_t=g_t^{\text{loc}}y^{\text{loc}}_t+g_t^{\text{ssm}}y^{\text{ssm}}_t+
g_t^{\text{mem}}y^{\text{mem}}_t,
\qquad h\leftarrow h+W_f r.
\label{eq:fusion}
\end{equation}
Because the state $(s,M)$ persists across segments, a window of only $w$ tokens can carry
dependencies far longer than $w$.

\paragraph{Programmable compute: the low--rank instruction bank.}
In place of a soft mixture of SwiGLU
experts~\cite{shazeer2017moe}, both projections of the SwiGLU are synthesized
per token from shared low--rank banks. Writing it for a generic projection $W$,
\begin{equation}
W(t) \;=\; W_{0} \;+\; \sum_{k=1}^{K} c_{t,k}\, U_k V_k^{\top},
\qquad c_t \in \mathbb{R}^{K},
\label{eq:bank}
\end{equation}
where $W_{0}$ is a shared base operator and $\{U_kV_k^{\top}\}$ are $K$ learned
rank--$r$ ``instructions,'' selected by the per--token code $c_t$. For the input map
$W_{\text{in}}:\mathbb{R}^{d}\!\to\!\mathbb{R}^{2d_{\text{ff}}}$ (the concatenated
gate/up projection) the bank has $V_k\in\mathbb{R}^{d\times r}$ and
$U_k\in\mathbb{R}^{2d_{\text{ff}}\times r}$; for the down map
$W_{\text{down}}:\mathbb{R}^{d_{\text{ff}}}\!\to\!\mathbb{R}^{d}$ a second bank has
$V^{d}_k\in\mathbb{R}^{d_{\text{ff}}\times r}$ and $U^{d}_k\in\mathbb{R}^{d\times r}$,
both gated by the same code $c_t$. Crucially, no token--specific weight is ever
materialized: each programmed projection is a base map plus a low--rank correction,
\begin{equation}
W_{\text{in}}(t)\,u_t \;=\; W_{\text{in},0}\,u_t
   \;+\; \sum_{k=1}^{K} c_{t,k}\,U_k\big(V_k^{\top}u_t\big),
\label{eq:apply}
\end{equation}
i.e.\ project $u_t$ onto the $K{\times}r$ bank, scale block $k$ by the code $c_{t,k}$,
and read out through $U$. Splitting this $2d_{\text{ff}}$--dimensional output into two
$d_{\text{ff}}$ halves $[\,\text{gate};\text{up}\,]=W_{\text{in}}(t)\,u_t$, the hidden
activation $h_t=\mathrm{SiLU}(\text{gate})\odot\text{up}$ is then mapped out through the
identically programmed $W_{\text{down}}(t)$. With $U_k,U^{d}_k$ initialized to zero
the layer starts exactly at the shared base SwiGLU. The per--token cost of the
programs is $\mathcal{O}\!\big(Kr(d{+}d_{\text{ff}})\big)$, so program capacity $K$
scales independently of the base FFN's $\mathcal{O}(d\,d_{\text{ff}})$
compute---each added instruction is $\sim\!r/d_{\text{ff}}$ as expensive as a full
MoE expert.

\paragraph{The instruction pointer: a manifold--conditioned code.}
The code is decoded from the token and a low--dimensional read--out of the carried SSM
state (Eq.~\ref{eq:ssm}),
\begin{equation}
c_t \;=\; \tanh\!\Big(\mathrm{MLP}\big([\,u_t\,;\, P\,y^{\text{ssm}}_t\,]\big)\Big)
\;\in\;[-1,1]^{K},
\qquad P\in\mathbb{R}^{m\times d},\ m\ll d,
\label{eq:code}
\end{equation}
so a slow, low--dimensional latent trajectory ($P\,y^{\text{ssm}}$) selects which
operator on the manifold executes at each token---the ``fetch--decode--execute''
loop, with the SSM state as program counter. The $\tanh$ keeps the synthesized
operator on a bounded region of the affine manifold.

\subsection{Memory carry and training}
Sequences are processed as contiguous segments of length $L$ with truncated
backpropagation through time: the state $(s,M)$ is threaded across segments and
detached every few segments; the Transformer runs the identical loop but is
stateless (full attention within each segment). The selective--SSM and delta--rule
memories use exact parallel forms (a log--depth associative scan and a chunkwise WY
triangular solve), so a training step vectorizes on GPU without per--timestep loops.

\section{Benchmarks and Setup}
\label{sec:benchmarks}
\paragraph{Joint neural codec (neural sequence generation and behavioral decoding).}
We use ``codec'' in the coder--decoder sense: from one shared hidden state the model
must both encode/continue the population spike code and decode behavior.
Concretely, each of three Neural Latents Benchmark~\cite{pei2021nlb} recordings---a
standard testbed for latent--dynamics models of motor cortex~\cite{pandarinath2018lfads}---is
turned into a dual--objective
problem: from the same hidden state the model must (a) autoregressively continue the
binned population spikes and (b) decode behavior---hand or finger velocity---through
a second linear head, minimizing
$\mathcal{L}=\mathrm{MSE}_{\text{next--spike}}+\lambda\,\mathrm{MSE}_{\text{decode}}$
($\lambda{=}10$, up--weighting the low--dimensional behavioral term against the
$N$--unit spike term). We use MC\_RTT (DANDI 000129, M1, random--target reach;
finger velocity, $N{=}130$ units), MC\_Maze (DANDI 000128, M1/PMd, maze
reaches; hand velocity, $N{=}182$), and Area2\_Bump (DANDI 000127,
somatosensory area~2; hand velocity, $N{=}65$). Spikes are binned at
\SI{50}{\milli\second}, smoothed with a Gaussian kernel over $3$ bins, and z--scored;
the neural continuation is seeded with a 128--bin prefix. We report the
behavioral--decoding $R^2$ on validation data, with a memoryless ridge decoder (short
causal history) as a reference. Neural next--step RMSE stays close to the
unit--variance noise floor (${\approx}1.0$) and is matched across architectures
throughout---single--trial spikes sit near that floor---so decode
$R^2$ is the comparison metric.

\paragraph{Setup.}
For each codec and architecture we build a ladder at fixed depth ($4$ layers)
and scale only width, targeting the same ${\sim}64$--$270$K non--embedding
parameter window for both architectures. Because VN--SST's carried state and instruction bank add a fixed
per--width overhead, matching this window puts the Transformer on wider blocks and
VN--SST on narrower ones, so the two curves overlap on the parameter axis rather
than occupying disjoint ranges; fixing depth additionally removes the shape confound
that otherwise injects large non--monotonicities into the Transformer's curve. The
parameter--scaling figure (Section~\ref{sec:param}) is run at a limited--data budget
(${\approx}25\%$ of the full training budget; ${\sim}2$/$14$/$7$K bins for
MC\_RTT/MC\_Maze/Area2, i.e.\ the smallest budget in Section~\ref{sec:data}), where models sit off the accuracy ceiling; the data-- and context--scaling sweeps use the
largest available training budget. All three neural sweeps report means over $3$ seeds; the
sequence--length sweep additionally holds the number of optimizer steps fixed across
context lengths (epochs scaled with $L$) so that context length is not confounded with
training budget. Both models use AdamW~\cite{loshchilov2019adamw} with a cosine
schedule~\cite{loshchilov2017sgdr} and warmup, gradient clipping, identical data
budgets and seeds, and $50$ training
epochs. VN--SST uses window $w$, SSM state
size $n$, memory dimensions $d_k{=}d_v$, a manifold read--out dimension $m$, and a
bank of $K$ rank--$r$ instructions as fixed hyperparameters (default $K{=}8$,
$r{=}8$, $m{=}8$). We do not match parameter counts; the
scaling line reveals efficiency directly.

\section{Results}
\label{sec:results}
We study VN--SST along the three axes a scaling analysis exposes---parameters, data,
and training context---always against a modern Transformer under matched budgets.
Throughout, the comparison metric is validation behavioral--decoding $R^2$;
single--trial spike prediction sits near its noise floor for both architectures
(Section~\ref{sec:benchmarks}), so it is the behavior read--out that separates the models.

\subsection{Parameter scaling under limited data}
\label{sec:param}
\begin{figure}[tbp]
\centering
\begin{subfigure}{0.32\textwidth}\includegraphics[width=\linewidth]{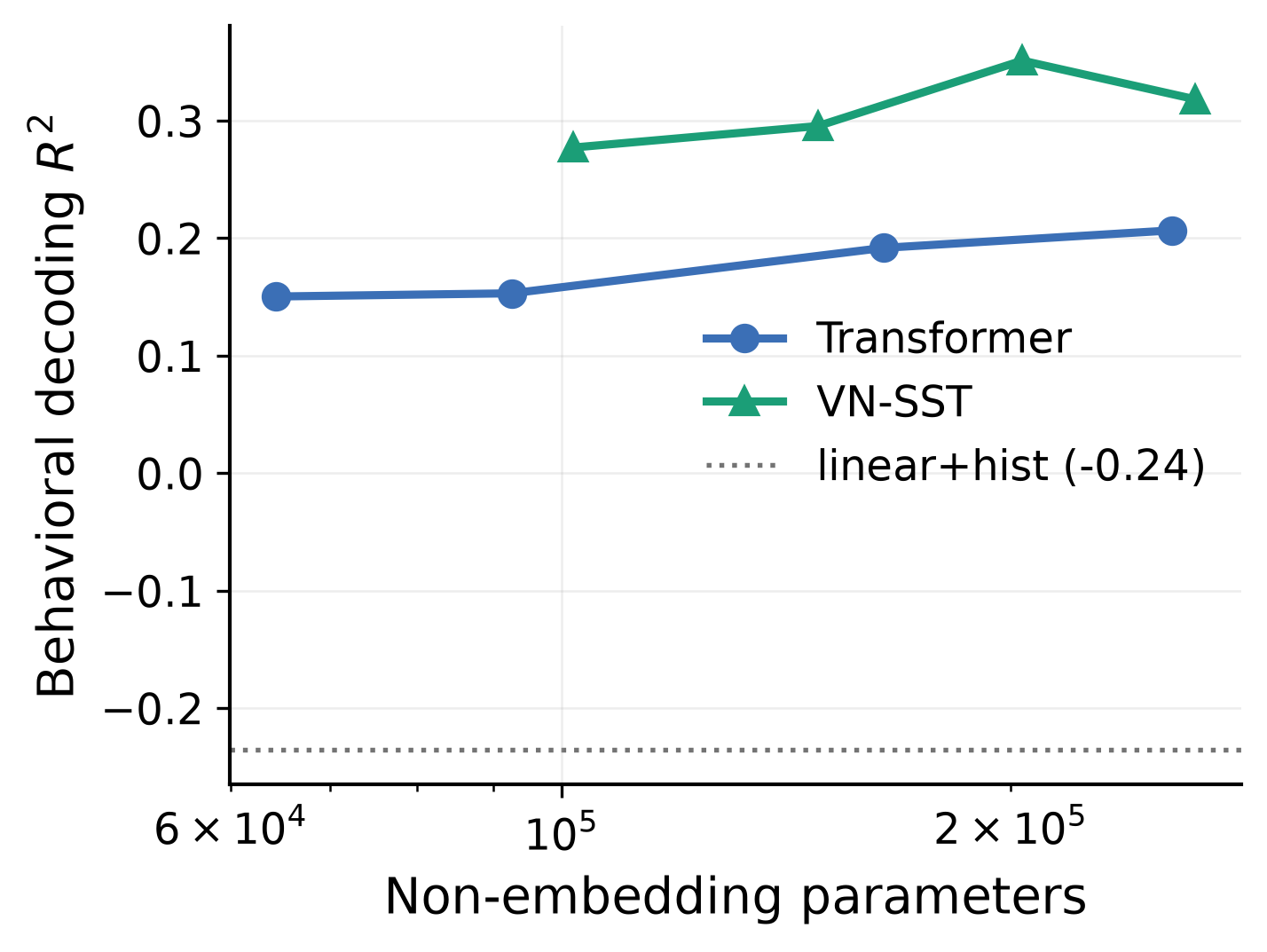}\caption{MC\_RTT (finger vel.)}\end{subfigure}\hfill
\begin{subfigure}{0.32\textwidth}\includegraphics[width=\linewidth]{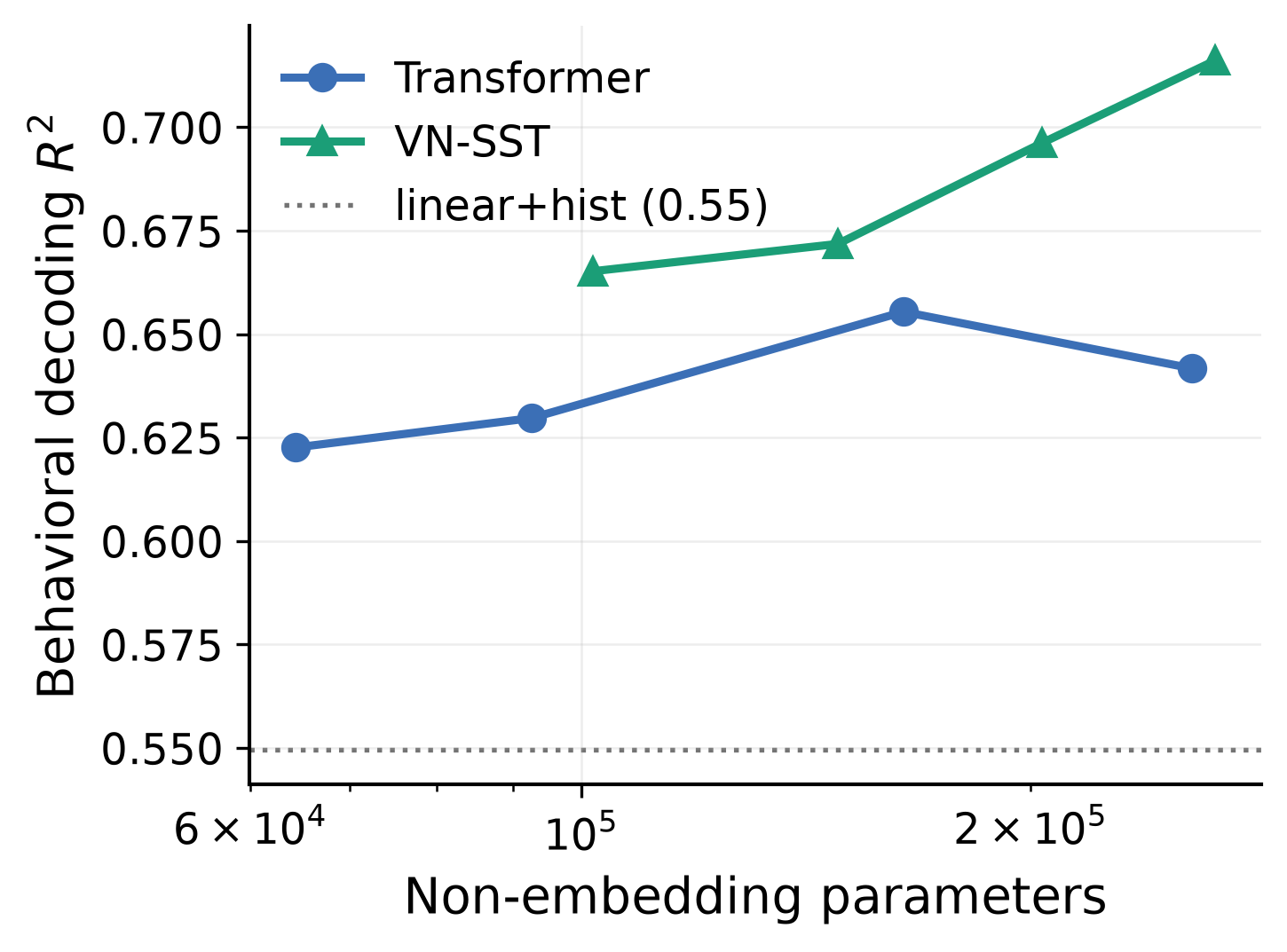}\caption{MC\_Maze (hand vel.)}\end{subfigure}\hfill
\begin{subfigure}{0.32\textwidth}\includegraphics[width=\linewidth]{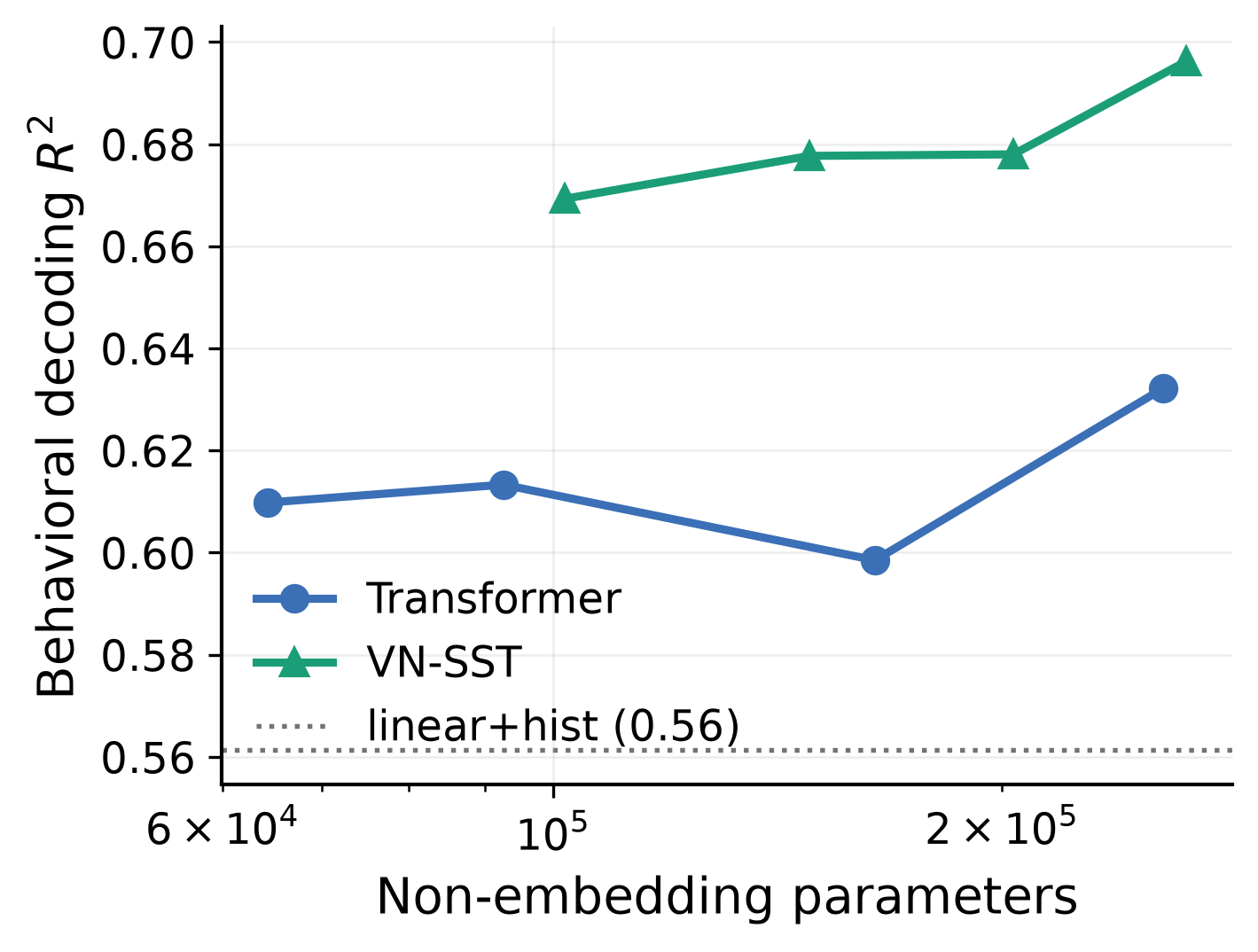}\caption{Area2\_Bump (hand vel.)}\end{subfigure}
\caption{Neural--codec parameter scaling with limited data. (a--c)~Behavioral--decoding
$R^2$ vs.\ non--embedding parameters on the MC\_RTT, MC\_Maze, and Area2\_Bump codecs,
respectively, for the Transformer (blue) and VN--SST (green, ``instruction bank''), with a
linear reference. The two ladders are aligned to the same ${\sim}64$--$270$K window
(Transformer on widths $40$--$80$, VN--SST on $24$--$48$), since VN--SST's carried state
and instruction bank add a fixed per--width overhead. Curves show the mean over $3$ seeds.
Away from the accuracy ceiling the architectures separate, with VN--SST leading on all three
codecs; the margin is largest on the data--scarce MC\_RTT codec in (a), where VN--SST reaches
$R^2{=}0.35$ against $0.21$ for the Transformer and the memoryless linear decoder is not
predictive ($R^2{<}0$).}
\label{fig:param}
\end{figure}

Table~\ref{tab:param} and Figure~\ref{fig:param} summarize the parameter sweep at a
limited--data budget (${\approx}25\%$ of the full training budget)---the scenario single--session
recordings actually occupy, and the one in which architecture matters most. Two things stand out. First, the parameter
ranges are aligned: because VN--SST's carried state and instruction bank add
a large fixed per--width overhead, matching the two architectures on raw parameters
requires the Transformer to use wider blocks (widths $40$--$80$) and VN--SST narrower ones
(widths $24$--$48$), so both curves cover the same ${\sim}64$--$270$K span rather than
sitting on disjoint ranges. Second, away from the full--data ceiling the curves
separate cleanly: replacing the dense FFN with the instruction bank helps most exactly
where data is scarce. On MC\_RTT peak decode $R^2$ is $0.351$ vs.\ $0.207$ (mean of $3$
seeds)---VN--SST beats the Transformer by a wide margin while the memoryless linear
decoder is not predictive ($R^2{=}{-}0.24$)---and it keeps a clear edge on the
better--sampled Area2\_Bump ($0.696$ vs.\ $0.632$) and MC\_Maze ($0.716$ vs.\ $0.655$)
codecs. Because the instruction bank synthesizes both SwiGLU projections per token, the gain
is representational---a richer per--token operator (Section~\ref{sec:model3})---rather than a
parameter effect, and unlike a dense FFN it comes with the measurable control channel
of Section~\ref{sec:controlbits}. Because each point is a $3$--seed mean and the
width dependence within a ladder is mild, the robust signal here is the separation
between architectures rather than the exact curvature of either curve.

\begin{table}[H]
\centering
\caption{Joint neural codec, parameter scaling with limited data: peak
behavioral--decoding $R^2$ (mean of $3$ seeds). ``Linear'' is a memoryless ridge
decoder with short causal history.}
\label{tab:param}
\small
\begin{tabular}{l l c c c}
\toprule
Benchmark & decode & Linear $R^2$ & Transformer $R^2$ & VN--SST $R^2$ \\
\midrule
MC\_RTT & finger vel. & -0.236 & 0.207 & 0.351 \\
MC\_Maze & hand vel. & 0.549 & 0.655 & 0.716 \\
Area2\_Bump & hand vel. & 0.561 & 0.632 & 0.696 \\
\bottomrule
\end{tabular}
\end{table}

\FloatBarrier
\subsection{Data scaling}
\label{sec:data}
We fix the model (the fixed--depth ${\sim}150$K decoding model) and vary the amount of
training data over four recording--bin budgets per codec.
\begin{table}[H]
\centering
\caption{Data scaling at the fixed--depth ${\sim}150$K decoding model ($50$ epochs,
smoothed rates, mean of $3$ seeds): decode $R^2$ at the smallest\,$\to$\,largest
training budget (four recording--bin budgets per codec).}
\label{tab:data}
\small
\begin{tabular}{l l cc}
\toprule
Benchmark & data ($n$) & Transformer & VN--SST \\
\midrule
MC\_RTT & 2k\,$\to$\,8k bins & 0.192\,$\to$\,0.522 & 0.295\,$\to$\,0.607 \\
MC\_Maze & 14k\,$\to$\,58k bins & 0.655\,$\to$\,0.847 & 0.672\,$\to$\,0.854 \\
Area2\_Bump & 7k\,$\to$\,28k bins & 0.599\,$\to$\,0.778 & 0.678\,$\to$\,0.795 \\
\bottomrule
\end{tabular}
\end{table}

\begin{figure}[tbp]
\centering
\begin{subfigure}{0.32\textwidth}\includegraphics[width=\linewidth]{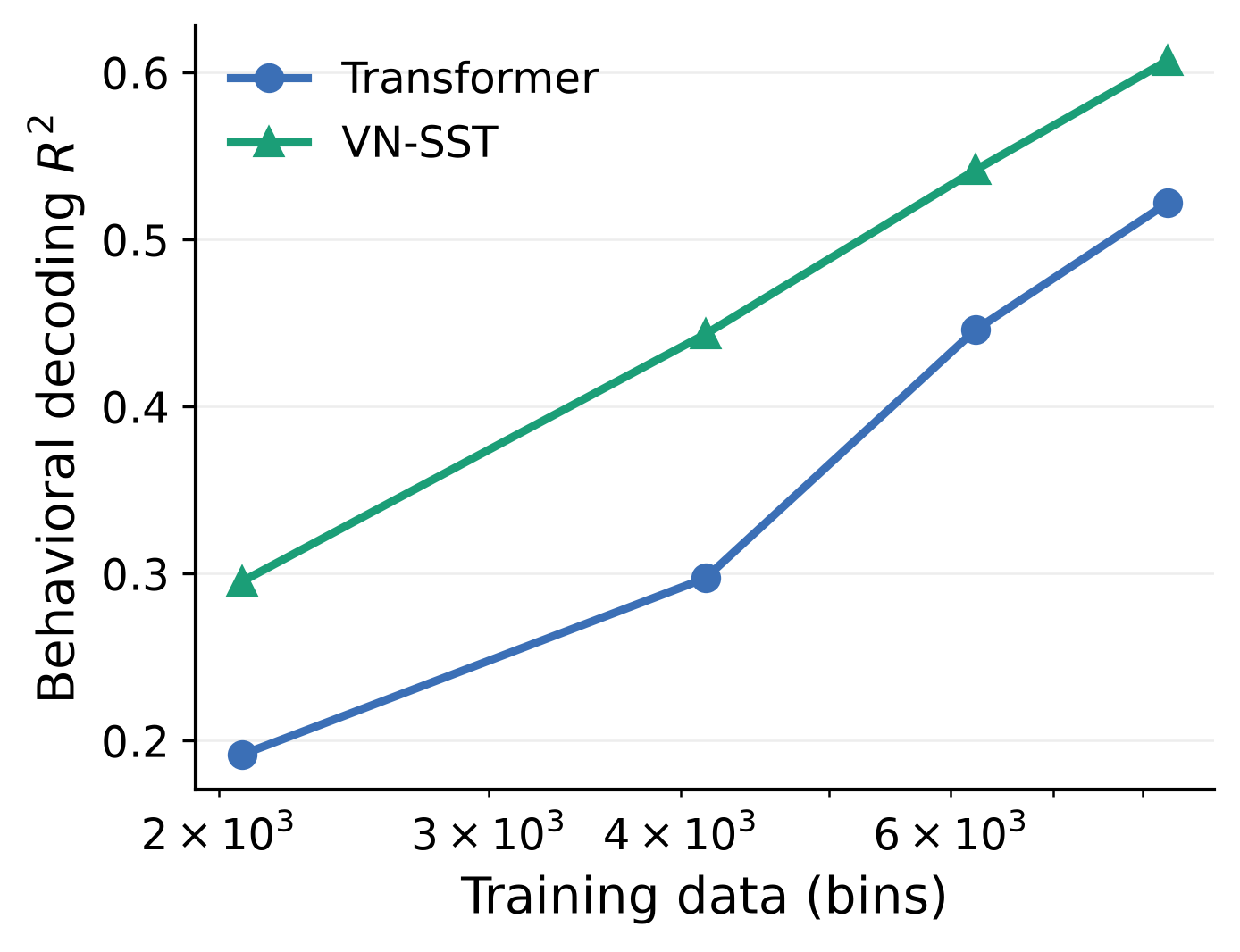}\caption{MC\_RTT codec}\end{subfigure}\hfill
\begin{subfigure}{0.32\textwidth}\includegraphics[width=\linewidth]{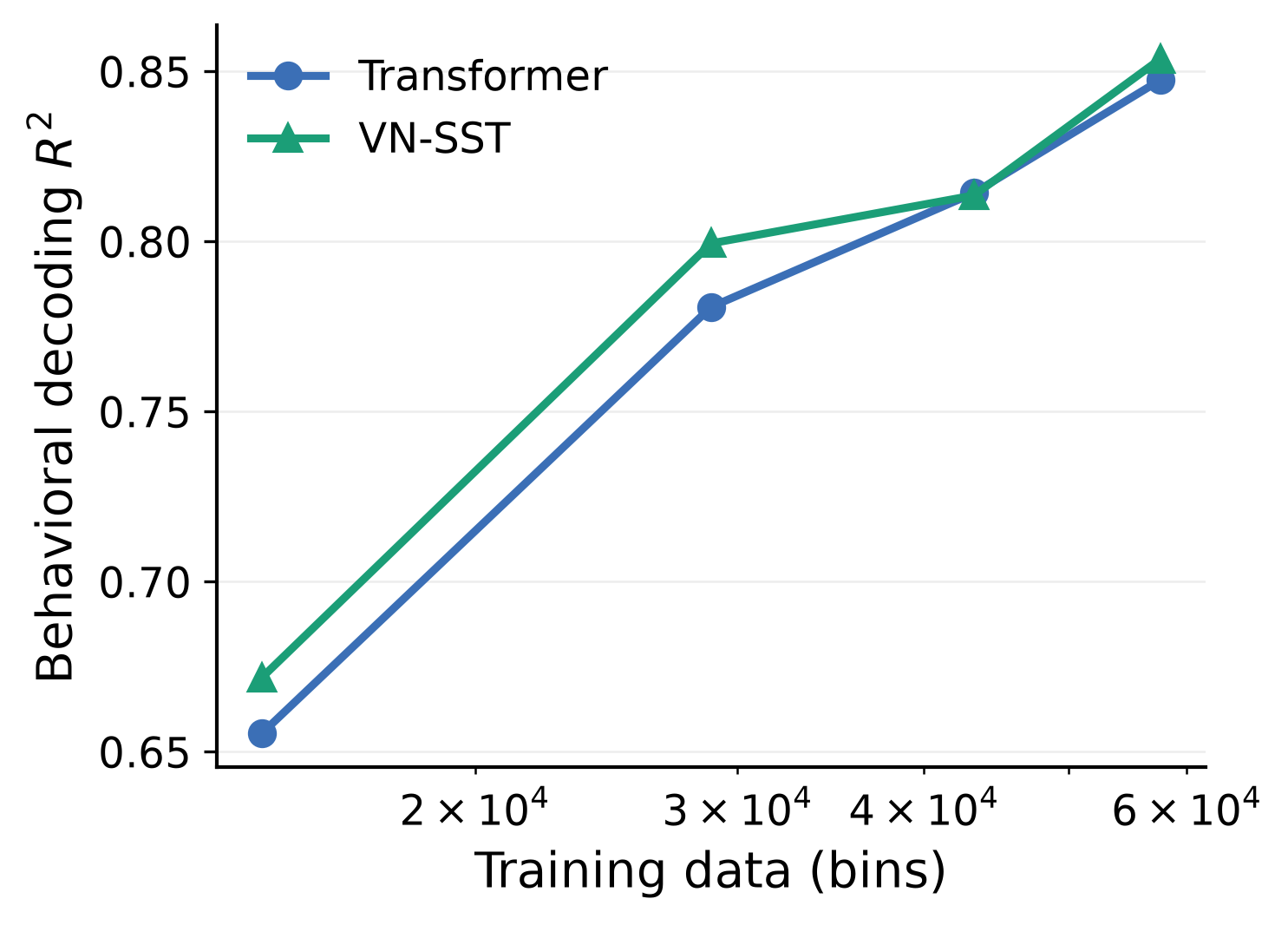}\caption{MC\_Maze codec}\end{subfigure}\hfill
\begin{subfigure}{0.32\textwidth}\includegraphics[width=\linewidth]{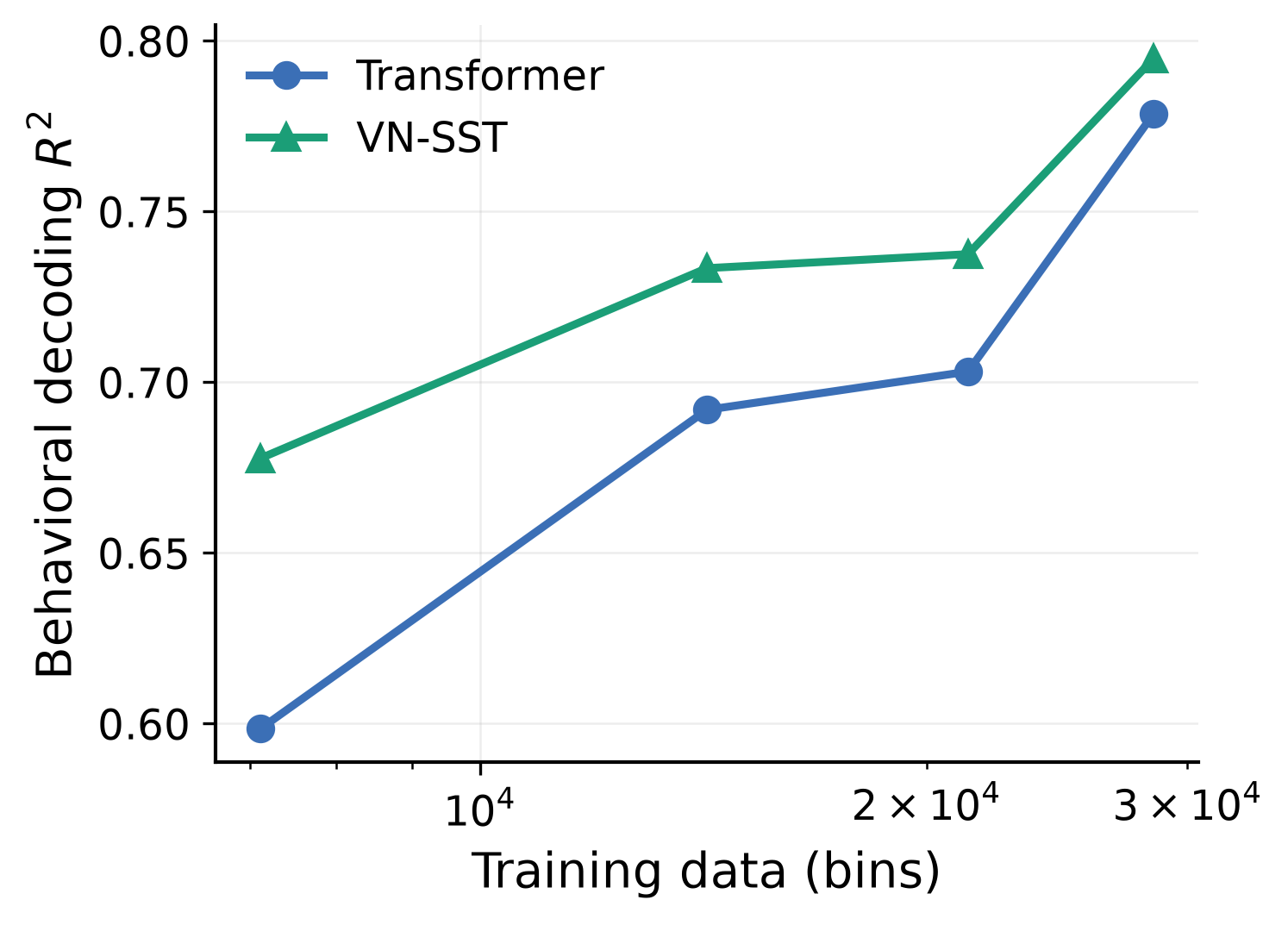}\caption{Area2\_Bump codec}\end{subfigure}
\caption{Data scaling at the fixed--depth ${\sim}150$K decoding model (mean of $3$ seeds).
(a--c)~Decode $R^2$ vs.\ training bins on the MC\_RTT, MC\_Maze, and Area2\_Bump codecs,
respectively, for the Transformer (blue) and VN--SST (green).}
\label{fig:data}
\end{figure}

Across the four data budgets VN--SST is at least as accurate as the Transformer on all
three codecs, and its advantage is largest where data is scarcest. At the smallest budget
it is ahead on Area2\_Bump (decode $R^2$ $0.68$ vs.\ $0.60$ at ${\sim}7$k bins) and on
MC\_RTT ($0.30$ vs.\ $0.19$ at ${\sim}2$k bins); the gap then narrows as data grows, with
the two models close at the full recording (MC\_RTT $0.61$ vs.\ $0.52$; MC\_Maze $0.85$
vs.\ $0.85$; Area2 $0.79$ vs.\ $0.78$). A plausible reading is that the carried
low--dimensional state supplies temporal context that partly substitutes for data, so the
clearest gains appear under the tightest budgets and diminish as data grows.

\FloatBarrier
\subsection{Sequence--length scaling}
\label{sec:seqlen}
Holding the model and the full training set fixed, we varied only the truncated--BPTT
segment length $L\in\{16,32,64,128\}$. Because the number of optimizer steps per epoch
scales as $T/L$, a fixed--epoch sweep would confound context length with training budget:
longer segments yield proportionally fewer updates. We therefore equalized the number of
optimizer steps across conditions by scaling the epoch count with $L$, and we report the
mean over $3$ seeds.
\begin{table}[H]
\centering
\caption{Sequence--length scaling at the fixed--depth ${\sim}150$K decoding model and
full data, at a matched training budget (epochs scaled with $L$ so the number
of optimizer steps is held constant across context lengths; mean of $3$ seeds):
best decode $R^2$ over $L\in\{16,32,64,128\}$ and the $L$ at which it peaks
(each cell reads best\,$R^2$\,@\,$L$). On the scarce MC\_RTT codec the two diverge---the Transformer's
decode gently declines as $L$ grows ($0.61\!\to\!0.57$) while VN--SST's rises
($0.60\!\to\!0.68$), converting longer context into accuracy through its carried
state; on the data--rich codecs both plateau (VN--SST slightly higher and flatter,
and it alone holds up at $L{=}128$).}
\label{tab:seqlen}
\small
\begin{tabular}{l l cc}
\toprule
Benchmark & seq.\ len.\ ($L$) & Transformer & VN--SST \\
\midrule
MC\_RTT & 16\,$\to$\,128 & 0.612\,@\,16 & 0.680\,@\,128 \\
MC\_Maze & 16\,$\to$\,128 & 0.879\,@\,32 & 0.878\,@\,16 \\
Area2\_Bump & 16\,$\to$\,128 & 0.766\,@\,32 & 0.779\,@\,64 \\
\bottomrule
\end{tabular}
\end{table}

\begin{figure}[tbp]
\centering
\begin{subfigure}{0.32\textwidth}\includegraphics[width=\linewidth]{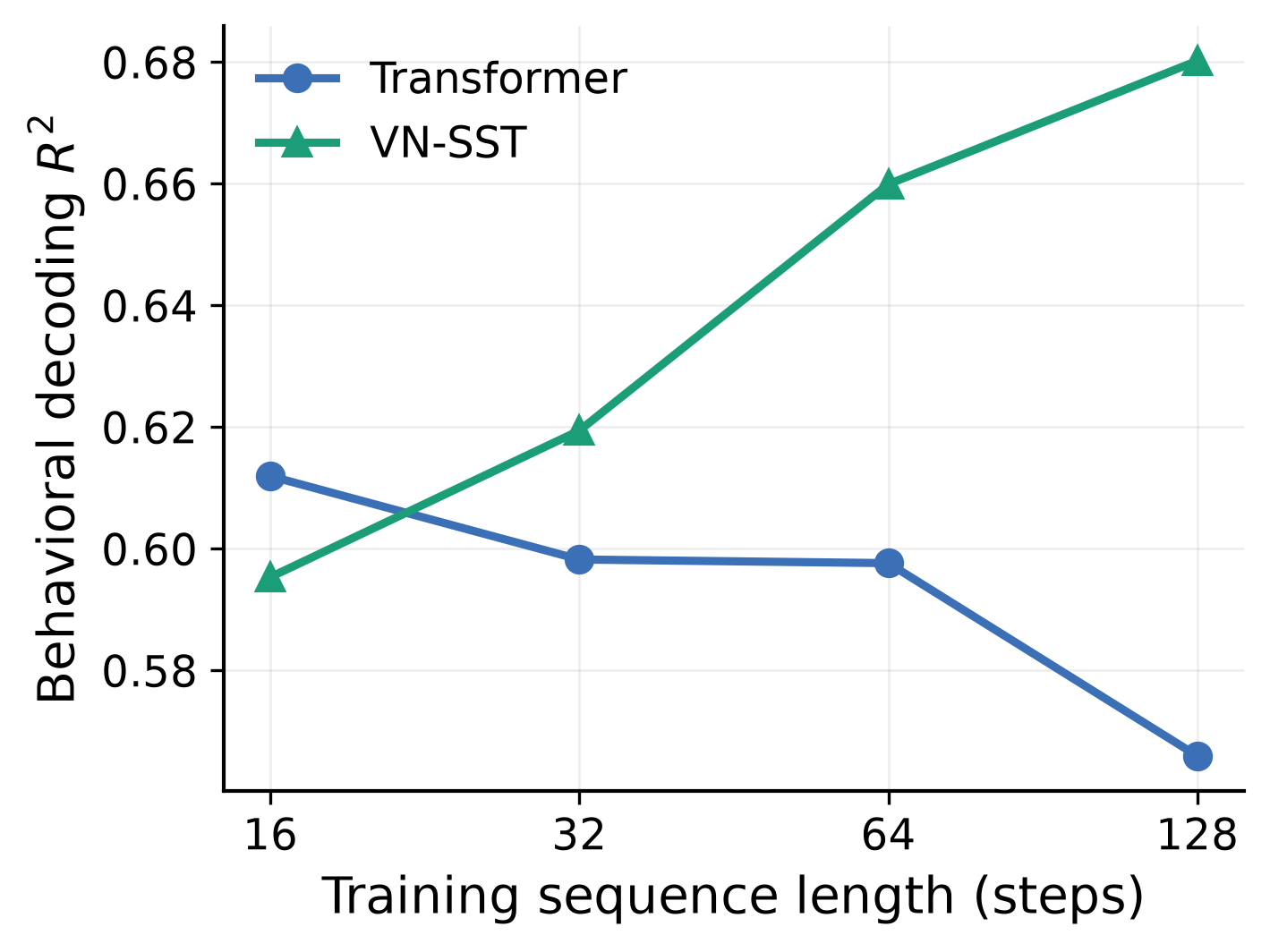}\caption{MC\_RTT codec}\end{subfigure}\hfill
\begin{subfigure}{0.32\textwidth}\includegraphics[width=\linewidth]{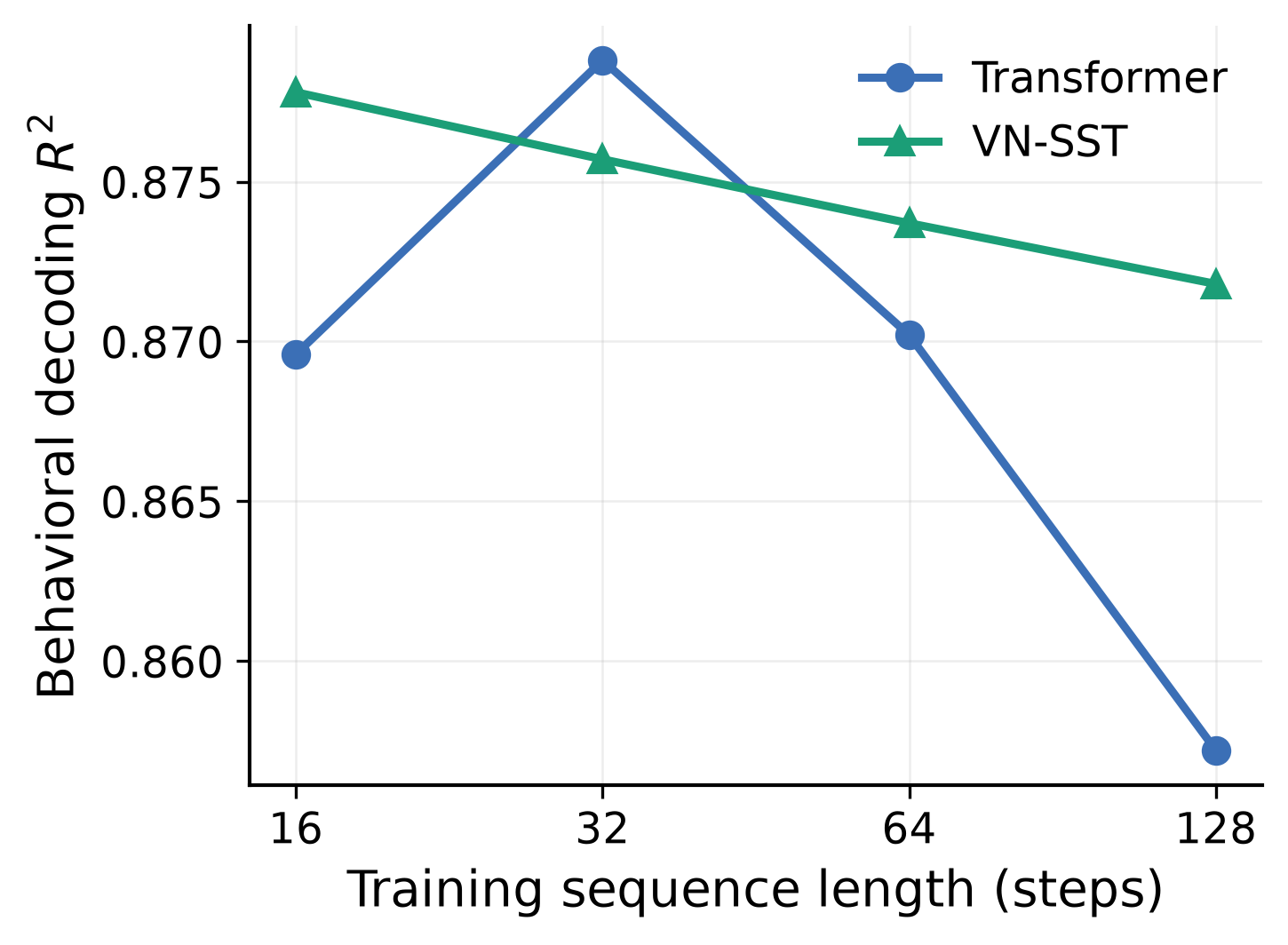}\caption{MC\_Maze codec}\end{subfigure}\hfill
\begin{subfigure}{0.32\textwidth}\includegraphics[width=\linewidth]{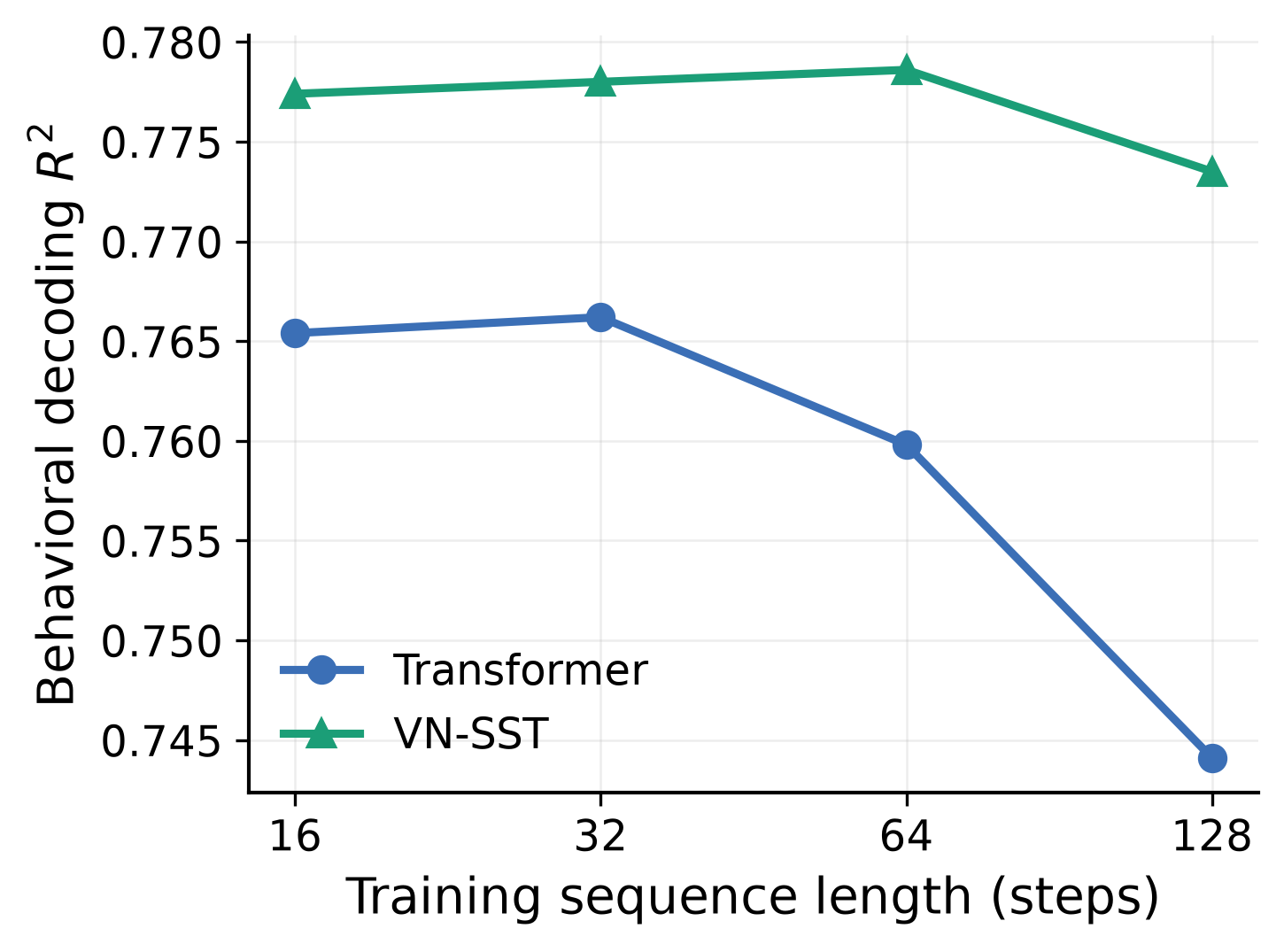}\caption{Area2\_Bump codec}\end{subfigure}
\caption{Sequence--length scaling at fixed model and full data, at a matched
update budget (epochs scaled with $L$; mean of $3$ seeds): decode $R^2$ vs.\ training
context length $L$ (log$_2$ $x$--axis), on the MC\_RTT (a), MC\_Maze (b), and
Area2\_Bump (c) codecs. (a)~On the scarce MC\_RTT codec the two architectures cross
over: the Transformer starts higher at $L{=}16$ but declines as the window grows
($0.61\!\to\!0.57$), whereas VN--SST rises ($0.60\!\to\!0.68$) and overtakes it---threading
state across segments, it converts longer context into decoding accuracy while the
Transformer's wider receptive field does not help its short--horizon behavioral read--out.
(b,c)~On the data--rich MC\_Maze and Area2\_Bump codecs both are high and largely flat,
with VN--SST slightly higher.}
\label{fig:seqlen}
\end{figure}

The context axis mirrors the data axis: longer context matters most where data is the
binding constraint, and it is where the two architectures behave most differently.
Once the update budget is matched---removing the fixed--epoch confound that would
otherwise exaggerate any trend---the scarce MC\_RTT codec shows a clean crossover. The
Transformer starts ahead at $L{=}16$ ($0.61$) but its decode drifts down as the window
grows. This is expected for the task rather than
a general context effect: instantaneous velocity is a short--horizon readout of the current
population state, so with the update budget already matched a longer window carries little
additional predictive signal. What it does add is a wider attention receptive field (more
capacity to overfit) and, since a fixed recording yields $T/L$ segments, fewer independent
training sequences per epoch---both of which slightly hurt generalization for a plain
Transformer, unlike language modeling, where the target genuinely depends on long--range
context. VN--SST instead rises monotonically and overtakes it by $L{\approx}24$. One reason
is likely that its carried low--dimensional state integrates the extra context for the
read--out without widening the local window. On the data--rich codecs both stay high and
largely flat (MC\_Maze ${\approx}0.87$, Area2 ${\approx}0.77$; Table~\ref{tab:seqlen}), but
VN--SST is consistently slightly higher.

\FloatBarrier
\subsection{The program manifold and control bits}
\label{sec:controlbits}
\begin{figure}[tbp]
\centering
\begin{subfigure}{0.32\textwidth}\includegraphics[width=\linewidth]{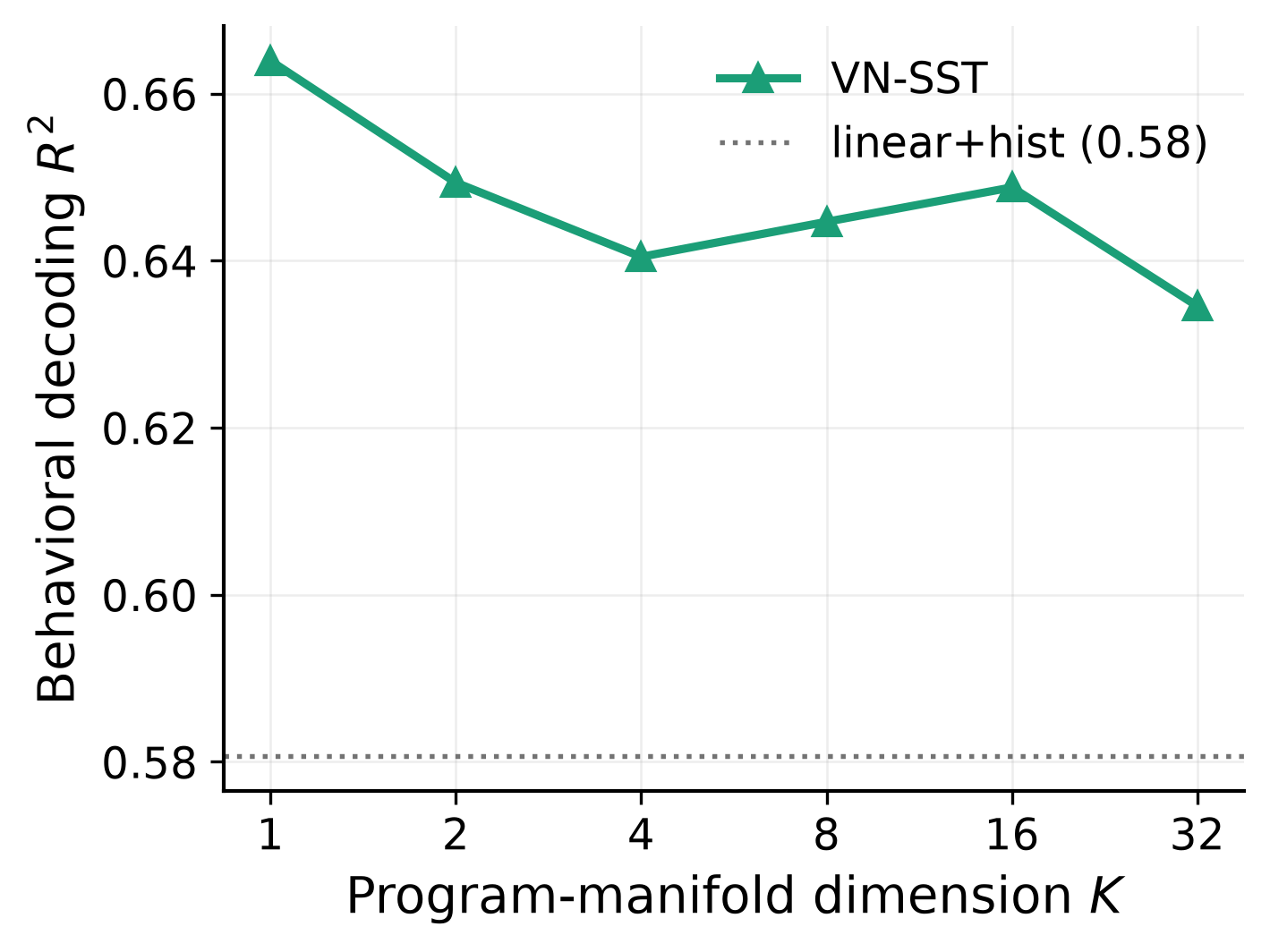}\caption{decode $R^2$ vs.\ $K$ (MC\_RTT)}\end{subfigure}\hfill
\begin{subfigure}{0.32\textwidth}\includegraphics[width=\linewidth]{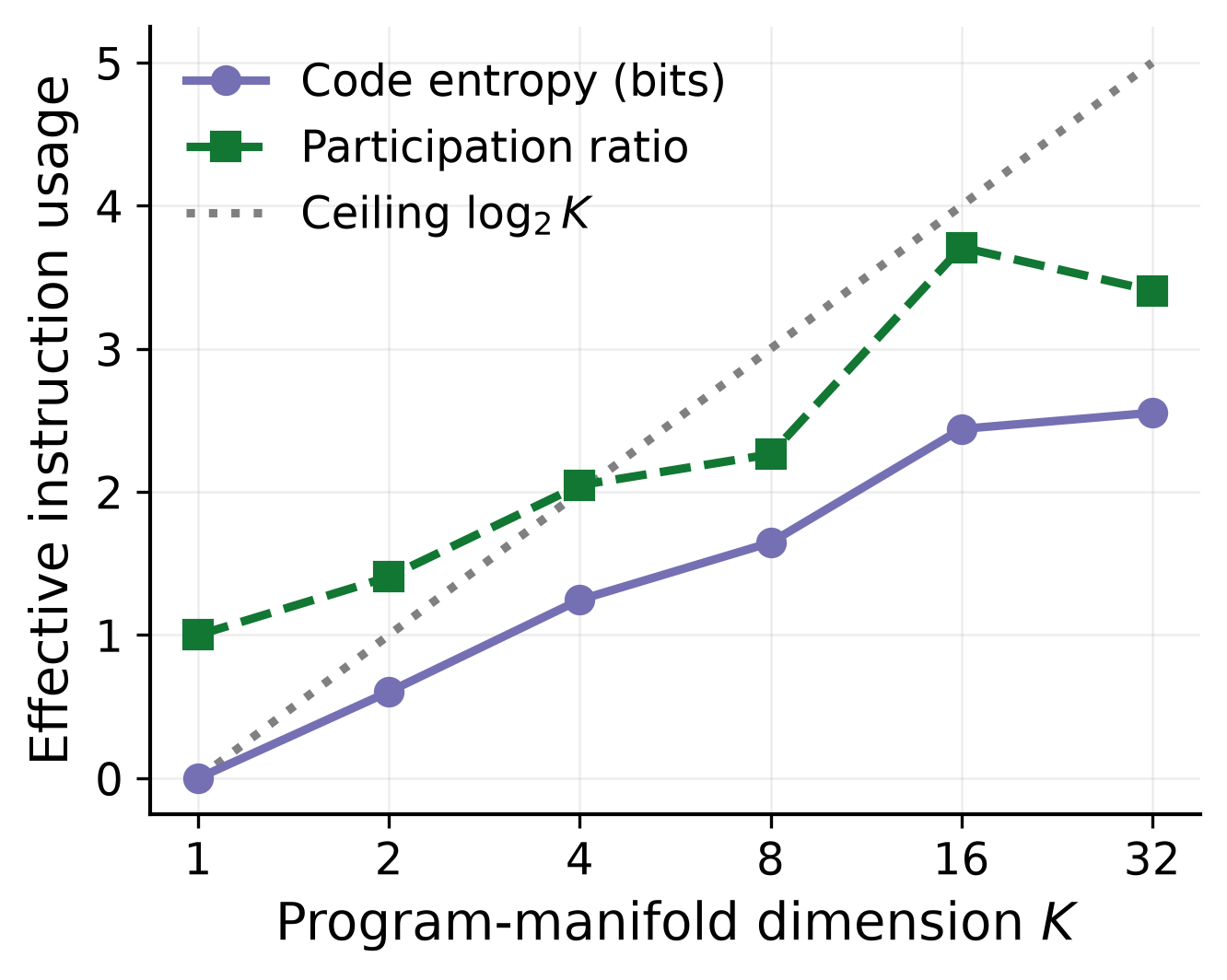}\caption{control bits (MC\_RTT)}\end{subfigure}\hfill
\begin{subfigure}{0.32\textwidth}\includegraphics[width=\linewidth]{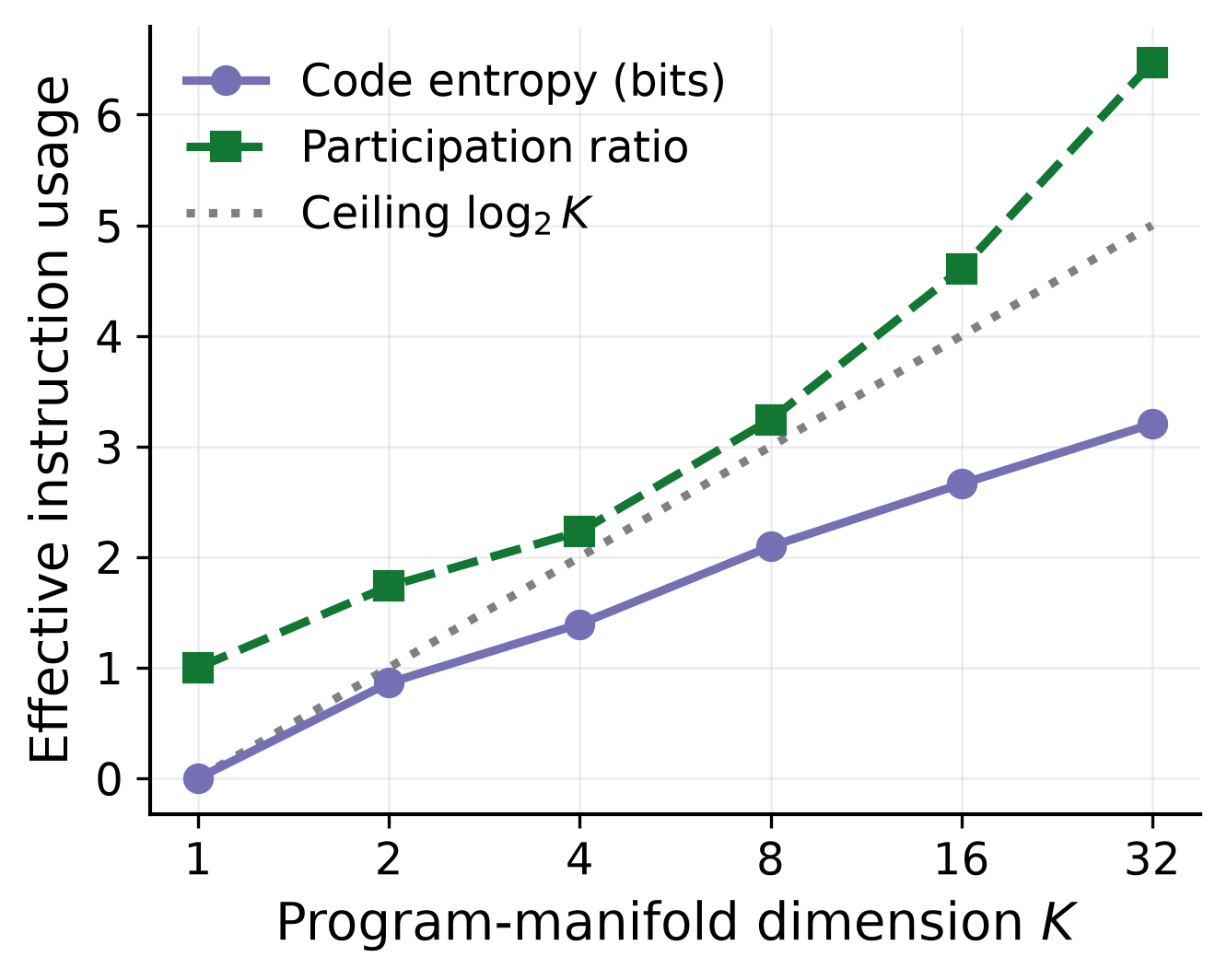}\caption{control bits (Area2)}\end{subfigure}
\caption{Program--manifold sweep at fixed model size. (a) Decode $R^2$ is essentially flat
and non--monotonic in the program count $K$---capacity is not the bottleneck here. (b,c)
The per--token instruction code compresses: given $K{=}32$ instructions the network
uses only ${\approx}2.5$--$3.2$ bits of code entropy and a participation ratio of
${\approx}3.4$--$6.5$, bending progressively below the $\log_2 K$ ceiling---so the network
uses only a few instructions, consistently across recordings.}
\label{fig:sweep}
\end{figure}

Unlike a dense FFN or an output--blending MoE, the instruction bank has an explicit
capacity knob---the program--manifold dimension $K$. At a fixed model size (so per--token compute
barely moves) we sweep $K\in\{1,\dots,32\}$ and, for every run, both measure decode
$R^2$ and collect the per--token codes $c_t$ on validation data to quantify how much
of the program space the network actually uses. We report two summaries of the
empirical code distribution: the spectral entropy of its covariance (``code bits,''
bits of instruction variation) and the participation ratio (the effective number of
instructions used). Table~\ref{tab:controlbits} and Figure~\ref{fig:sweep} report the
result.
\begin{table}[H]
\centering
\caption{Control--bits diagnostic. Given a program bank of
size $K{=}32$ at a fixed model size, we collect the per--token instruction codes
$c_t$ on validation data and report the spectral entropy of their covariance (bits of
code variation) and the participation ratio (effective number of instructions used),
alongside the range of decode $R^2$ over $K\in\{1,\dots,32\}$. On every codec the
model compresses a $32$--instruction bank to ${\approx}2.5$--$3.2$ bits /
${\approx}3.4$--$6.5$ effective operators---well below the $\log_2 K{=}5$ ceiling---while
decode $R^2$ is essentially flat in $K$: program capacity is a control/compression
knob, not an accuracy lever.}
\label{tab:controlbits}
\small
\begin{tabular}{l c c c c c}
\toprule
Benchmark & $K$ & code bits & $\log_2 K$ & particip.\ ratio & decode $R^2$ (range) \\
\midrule
MC\_RTT & 32 & 2.55 & 5.0 & 3.4 & 0.635--0.664 \\
MC\_Maze & 32 & 2.89 & 5.0 & 5.1 & 0.864--0.872 \\
Area2\_Bump & 32 & 3.20 & 5.0 & 6.5 & 0.779--0.787 \\
\bottomrule
\end{tabular}
\end{table}

In this study, two observations emerge, and they are the point of the architecture. (1)~Program
capacity is not an accuracy lever here. Decode $R^2$ is essentially flat and
non--monotonic across $K$ (Fig.~\ref{fig:sweep}a): the small motor codes these
recordings support are already captured by a handful of operators, so adding
instructions neither helps nor hurts. (2)~The network compresses to a small
instruction set. Given a bank of $K{=}32$, the empirical code uses only
${\approx}2.55$ bits (MC\_RTT), $3.20$ (Area2), and $2.89$ (MC\_Maze)---against a
ceiling of $\log_2 32=5$---with participation ratios of ${\approx}3.4$--$6.5$ effective
instructions (Table~\ref{tab:controlbits}). The entropy curve bends below the ceiling
as $K$ grows (Fig.~\ref{fig:sweep}b,c): the model does not spread across all available
programs but concentrates on a few, and it does so consistently across three brain
areas. This is the von--Neumann claim made numeric---the layer runs on ${\approx}2.5$--$3$
bits of program per token---and it is a property only operator synthesis can
report. An output--blending MoE has no addressable instruction whose entropy can be
measured. It also suggests a design rule and an interpretability
handle (which few operators specialize, and how they compose) that we leave to future
work.

\FloatBarrier
\subsection{Additional modality check: language}
\label{sec:lm}
Although this paper is about neural decoding, the instruction bank is a generic
feed--forward mechanism, so it is worth asking whether it also helps on a very
different sequence modality. We therefore additionally tested the same
VN--SST (identical bank size $K$, rank $r$, and manifold read--out) against the
Transformer on two sub--word (byte--level BPE) text corpora,
tiny--Shakespeare and WikiText--2, over a fixed--depth, width--scaled ladder (two layers;
${\sim}0.13$--$1.7$M non--embedding parameters; mean of $3$ seeds), matching the
fixed--depth protocol used for the neural codecs. To see how the comparison moves with
data, we run each corpus at two budgets: a small ``1$\times$'' budget and a ``3$\times$''
budget with three times as many training and validation tokens (tiny--Shakespeare
$114$K${\to}342$K, its full corpus; WikiText--2 $440$K${\to}1.32$M).

\begin{figure}[tbp]
\centering
\begin{subfigure}{0.48\textwidth}\includegraphics[width=\linewidth]{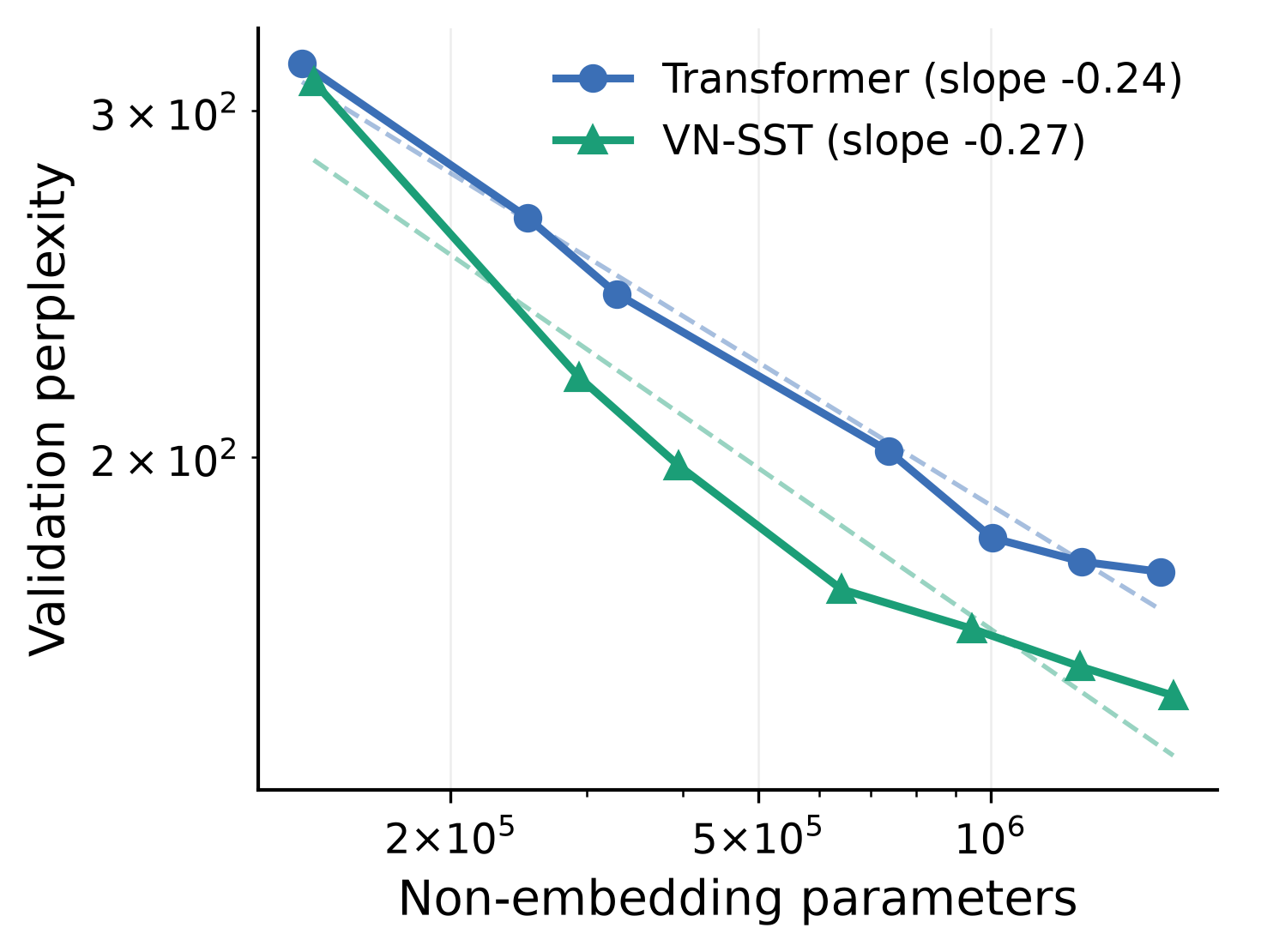}\caption{tiny--Shakespeare, 1$\times$ ($114$K tokens)}\end{subfigure}\hfill
\begin{subfigure}{0.48\textwidth}\includegraphics[width=\linewidth]{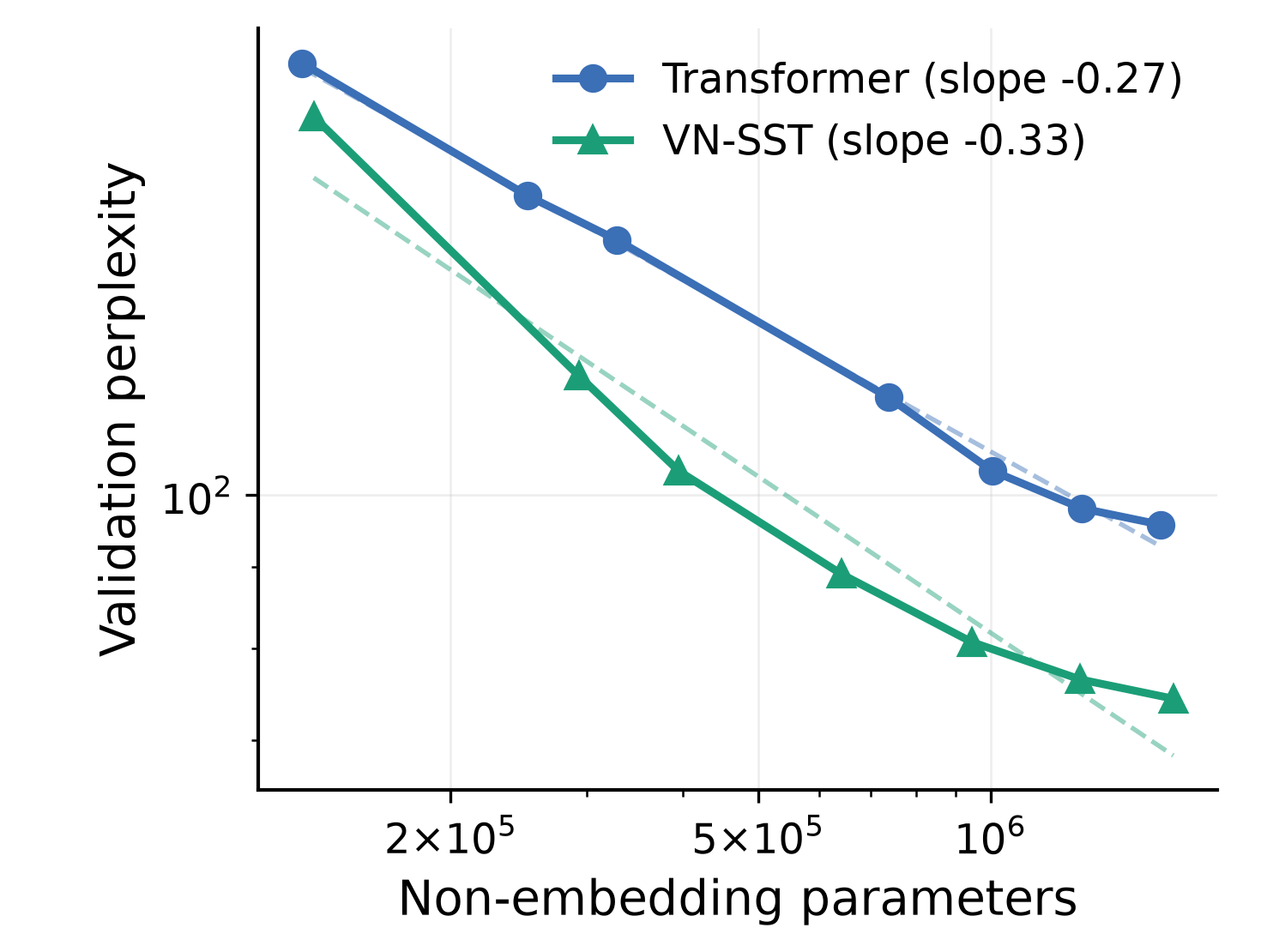}\caption{WikiText--2, 1$\times$ ($440$K tokens)}\end{subfigure}

\vspace{4pt}
\begin{subfigure}{0.48\textwidth}\includegraphics[width=\linewidth]{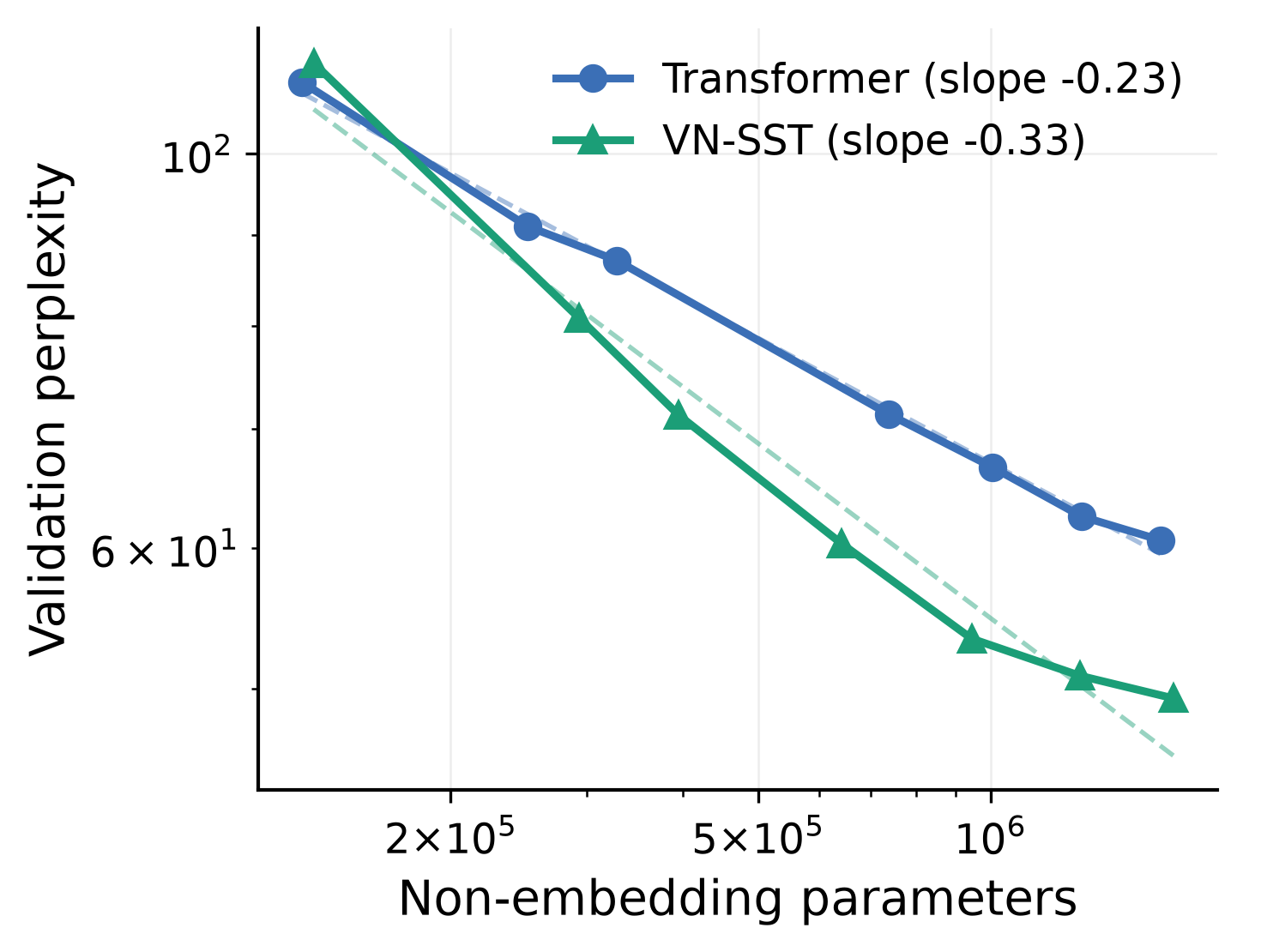}\caption{tiny--Shakespeare, 3$\times$ ($342$K tokens)}\end{subfigure}\hfill
\begin{subfigure}{0.48\textwidth}\includegraphics[width=\linewidth]{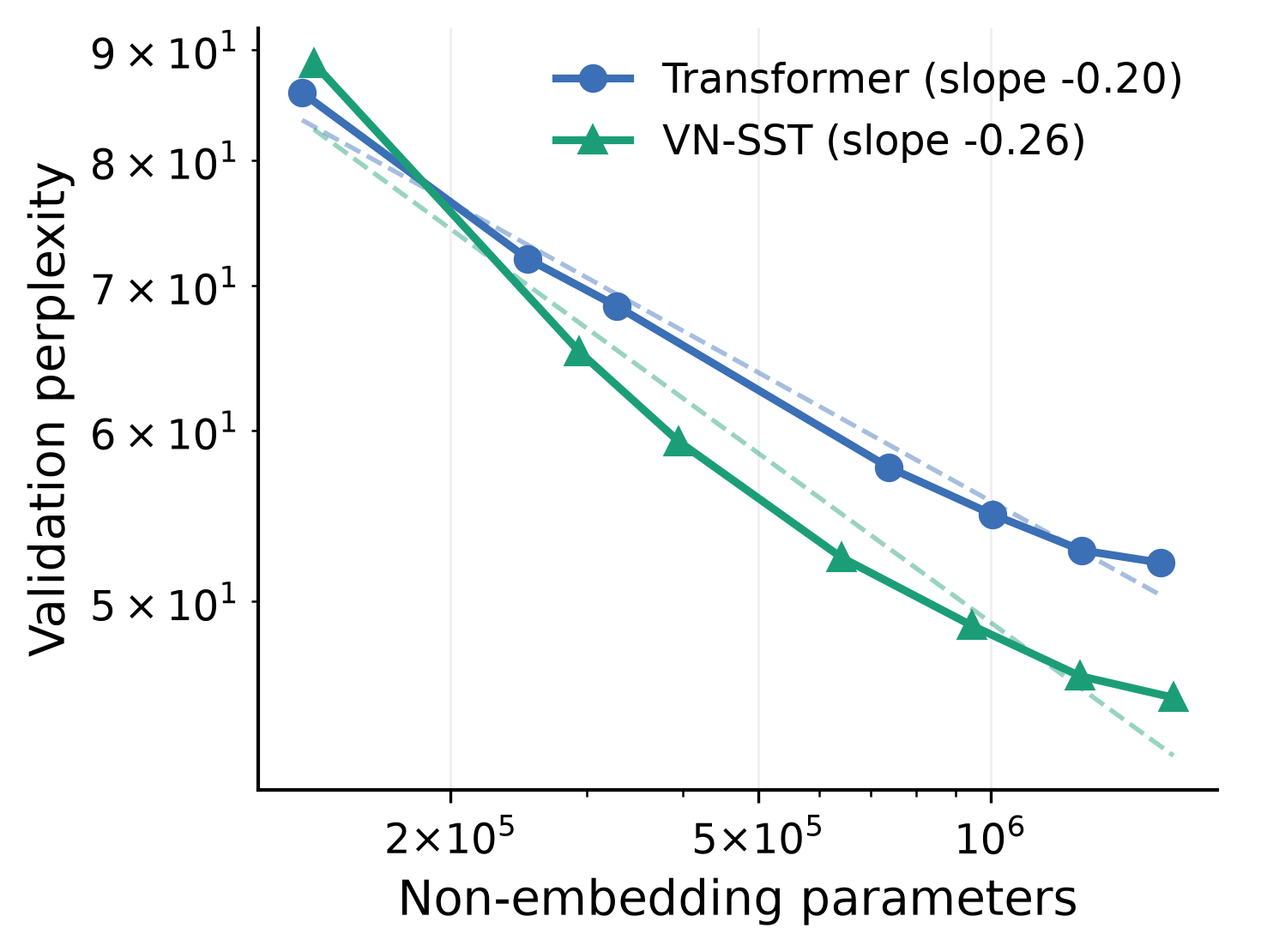}\caption{WikiText--2, 3$\times$ ($1.32$M tokens)}\end{subfigure}
\caption{Additional modality check: language--model scaling at two data budgets. Validation
perplexity vs.\ non--embedding parameters (log--log, with power--law fits) for the Transformer
and VN--SST. Top row (a,b): the small ``1$\times$'' budget; bottom row (c,d): the ``3$\times$''
budget with three times as many tokens; tiny--Shakespeare in (a,c) and WikiText--2 in (b,d).
The instruction--bank model---unchanged from the neural experiments---tracks below the
Transformer across the ladder in every panel, has the steeper slope, and reaches the best
perplexity. More data lowers both curves; VN--SST's lead widens with data on
tiny--Shakespeare and narrows on WikiText--2 (where both models saturate), but it stays ahead
throughout.}
\label{fig:lm}
\end{figure}

Table~\ref{tab:lm} and Figure~\ref{fig:lm} show that the instruction bank transfers, and
that the effect is robust to data scale. At both budgets and on both corpora, VN--SST tracks
below the Transformer across the ladder and has the steeper power--law slope, so it is more
parameter--efficient throughout. Adding data helps both models---perplexity falls sharply from
the 1$\times$ to the 3$\times$ budget (WikiText--2 top--of--ladder $95.7\!\to\!52.1$ for the
Transformer and $74.4\!\to\!45.1$ for VN--SST; tiny--Shakespeare $174.8\!\to\!60.6$ and
$151.2\!\to\!49.4$)---but it does not erase the gap. At the largest budget VN--SST still attains
the best perplexity on each corpus, $49.4$ vs.\ $60.6$ on tiny--Shakespeare and $45.1$ vs.\
$52.1$ on WikiText--2 (a ${\approx}13$--$18\%$ reduction at the top of the ladder), and it reaches
the Transformer's best with roughly $2$--$3\times$ fewer parameters. The one nuance is how the
lead moves with scale: on tiny--Shakespeare it widens with more data (VN--SST's slope steepens
to $-0.33$), whereas on WikiText--2 the slopes flatten as both models saturate and the margin
narrows, though VN--SST stays ahead. We read this as
evidence that per--token operator synthesis is a generic gain---a programmable operator
plus the state and fast--weight pathways help wherever long--range structure must be
integrated from a small window---rather than a quirk of the neural codecs. We keep the
study framed
around neural decoding, where the small--data, small--model regime makes the memory
and program machinery most decisive, and treat language as a check on generality.

\begin{table}[H]
\centering
\caption{Additional modality check---sub--word language modeling. Validation
perplexity (best over a ${\sim}0.13$--$1.7$M non--embedding parameter ladder at fixed
two--layer depth with width scaled, mean of $3$ seeds, at the largest ($3\times$) data
budget) and the fitted power--law slope vs.\ parameters, for the
Transformer and VN--SST on byte--level BPE tiny--Shakespeare and WikiText--2. The
instruction--bank model, unchanged from the neural experiments, is more parameter--efficient
across the ladder on both corpora, attains the best perplexity on both, and has the
steeper slope on both.}
\label{tab:lm}
\small
\begin{tabular}{l cc cc}
\toprule
& \multicolumn{2}{c}{Transformer} & \multicolumn{2}{c}{VN--SST} \\
\cmidrule(lr){2-3}\cmidrule(lr){4-5}
Corpus & best PPL & slope & best PPL & slope \\
\midrule
tiny--Shakespeare & 61 & -0.23 & 49 & -0.33 \\
WikiText--2 & 52 & -0.20 & 45 & -0.26 \\
\bottomrule
\end{tabular}
\end{table}

\FloatBarrier
\section{Conclusion}
We replaced a Transformer's fixed feed--forward operator with a low--rank instruction
bank---a shared base operator plus $K$ rank--$r$ instructions---from which a per--token,
manifold--conditioned code synthesizes the operator that the layer actually executes. The code is read
from a carried low--dimensional state, so a slow latent trajectory---the signal
that organizes cortical population activity---acts as an instruction pointer.
This makes the von--Neumann ``programmable FFN'' concrete: rather than blending the
outputs of a few fixed experts, the layer constructs a token--specific operator on a
low--dimensional weight manifold.

The mechanism earns its keep exactly where real neural recordings live---small data and
small models. On three motor--cortex decoding codecs VN--SST matches or beats a modern
Transformer along all three scaling axes: under a limited--data budget it leads by a
wide margin on the scarcest codec,
it is at least as accurate at every data budget on all three, and---uniquely---it turns a longer context
window into rising rather than falling decode accuracy. And because the operator is
addressed rather than blended, the control signal itself is measurable: the network
compresses a $32$--instruction bank to only ${\approx}2.5$--$3.2$ bits
(${\approx}3.4$--$6.5$ effective operators) per token, while decode accuracy stays flat
in $K$. Program capacity is therefore a control channel, not an accuracy lever.

More broadly, these results suggest that when data and models are scarce, the useful
inductive bias is not simply more parameters but a programmable operator steered by
low--dimensional dynamics---computation that, like the cortex it models, runs on a few
well--chosen instructions. Two pieces are reusable beyond neural decoding: a
parameter--efficient way to make feed--forward compute token--programmable, and a way to
quantify how much program a trained model actually uses.

\section*{Code and data availability}
Code to reproduce all experiments, figures, and tables---including the model
implementations, training loop, and scaling sweeps---is available from the authors upon
request, subject to internal review.
The neural benchmarks are the public Neural Latents Benchmark~\cite{pei2021nlb}
recordings on DANDI (dandisets 000127, 000128, and 000129); the text corpora are
tiny--Shakespeare and WikiText--2, both publicly available.

\section*{Acknowledgments}
We are grateful for our institution's support of this study. We thank the Neural Latents
Benchmark and DANDI teams for curating and publicly hosting the datasets used in this work.
This work is intended for scholarly purposes.


\begin{thebibliography}{99}
\bibitem{shazeer2017moe} N.\ Shazeer, A.\ Mirhoseini, K.\ Maziarz, A.\ Davis, Q.\ Le,
G.\ Hinton, and J.\ Dean. Outrageously large neural networks: The sparsely--gated
mixture--of--experts layer. \emph{International Conference on Learning Representations
(ICLR)}, 2017. arXiv:1701.06538.
\bibitem{kaplan2020scaling} J.\ Kaplan, S.\ McCandlish, T.\ Henighan, T.\ B.\ Brown,
B.\ Chess, R.\ Child, S.\ Gray, A.\ Radford, J.\ Wu, and D.\ Amodei. Scaling laws for
neural language models. \emph{arXiv:2001.08361}, 2020.
\bibitem{vaswani2017} A.\ Vaswani, N.\ Shazeer, N.\ Parmar, J.\ Uszkoreit, L.\ Jones,
A.\ N.\ Gomez, {\L}.\ Kaiser, and I.\ Polosukhin. Attention is all you need.
\emph{Advances in Neural Information Processing Systems (NeurIPS)}, 2017.
arXiv:1706.03762.
\bibitem{touvron2023llama} H.\ Touvron, T.\ Lavril, G.\ Izacard, X.\ Martinet, M.-A.\ Lachaux,
T.\ Lacroix, B.\ Rozi{\`e}re, N.\ Goyal, E.\ Hambro, F.\ Azhar, A.\ Rodriguez, A.\ Joulin,
E.\ Grave, and G.\ Lample. LLaMA: Open and efficient foundation language models.
\emph{arXiv:2302.13971}, 2023.
\bibitem{zhang2019rmsnorm} B.\ Zhang and R.\ Sennrich. Root mean square layer
normalization. \emph{Advances in Neural Information Processing Systems (NeurIPS)}, 2019.
arXiv:1910.07467.
\bibitem{su2021roformer} J.\ Su, Y.\ Lu, S.\ Pan, B.\ Wen, and Y.\ Liu. RoFormer:
Enhanced transformer with rotary position embedding. \emph{arXiv:2104.09864}, 2021.
\bibitem{shazeer2020glu} N.\ Shazeer. GLU variants improve transformer.
\emph{arXiv:2002.05202}, 2020.
\bibitem{gu2022s4} A.\ Gu, K.\ Goel, and C.\ R{\'e}. Efficiently modeling long sequences
with structured state spaces. \emph{International Conference on Learning Representations
(ICLR)}, 2022. arXiv:2111.00396.
\bibitem{gu2023mamba} A.\ Gu and T.\ Dao. Mamba: Linear--time sequence modeling with
selective state spaces. \emph{arXiv:2312.00752}, 2023.
\bibitem{pei2021nlb} F.\ Pei, J.\ Ye, D.\ Zoltowski, A.\ Wu, R.\ H.\ Chowdhury,
H.\ Sohn, J.\ E.\ O'Doherty, K.\ V.\ Shenoy, M.\ T.\ Kaufman, M.\ Churchland,
M.\ Jazayeri, L.\ E.\ Miller, J.\ Pillow, I.\ M.\ Park, E.\ L.\ Dyer, and C.\ Pandarinath.
Neural Latents Benchmark '21: Evaluating latent variable models of neural population
activity. \emph{Advances in Neural Information Processing Systems (NeurIPS) Datasets and
Benchmarks Track}, 2021. arXiv:2109.04463.
\bibitem{pandarinath2018lfads} C.\ Pandarinath, D.\ J.\ O'Shea, J.\ Collins, R.\ Jozefowicz,
S.\ D.\ Stavisky, J.\ C.\ Kao, E.\ M.\ Trautmann, M.\ T.\ Kaufman, S.\ I.\ Ryu,
L.\ R.\ Hochberg, J.\ M.\ Henderson, K.\ V.\ Shenoy, L.\ F.\ Abbott, and D.\ Sussillo.
Inferring single--trial neural population dynamics using sequential auto--encoders.
\emph{Nature Methods}, 15:805--815, 2018. doi:10.1038/s41592--018--0109--9.
\bibitem{loshchilov2019adamw} I.\ Loshchilov and F.\ Hutter. Decoupled weight decay
regularization. \emph{International Conference on Learning Representations (ICLR)}, 2019.
arXiv:1711.05101.
\bibitem{loshchilov2017sgdr} I.\ Loshchilov and F.\ Hutter. SGDR: Stochastic gradient
descent with warm restarts. \emph{International Conference on Learning Representations
(ICLR)}, 2017. arXiv:1608.03983.
\end{thebibliography}
\end{document}